# Distances with Mixed-Type Variables, some Modified Gower's Coefficients

*Marcello D'Orazio*
*Italian national Institute of Statistics-Istat, Rome, Italy*
*marcello.dorazio@istat.it*

**Abstract**

Nearest-neighbor methods have become popular in official statistics, mainly in imputation or in statistical matching problems; they play a key role in machine learning too, where a high number of variants have been proposed. Key decisions in nearest-neighbor methods concern the variables to use (when many potential candidates exist) and the distance function to apply. The first decision depends on the scope of the application and will not be the tackled in this article. The choice of the distance function depends mainly on the type of the selected variables. Unfortunately, relatively few options permit to handle mixed-type variables, a situation frequently encountered in official statistics.

The most popular distance for mixed-type variables is derived as the complement of the Gower's similarity coefficient; it is appealing because ranges between 0 and 1, being an average of the scaled distances calculated variable by variable. In addition, the Gower's distance handles missing values and allows for a user-defined weighting scheme when averaging distances. The discussion on the weighting schemes is sometimes misleading since it often ignores that in the unweighted "standard" setting the contribution of the single variables to the overall Gower's distance is unbalanced because of the different nature of the variables themselves.

This article tries to address the main drawbacks that affect the overall unweighted Gower's distance by suggesting some modifications in calculating the distance on the interval and ratio scaled variables. Simple modifications try to attenuate the impact of outliers on the scaled Manhattan distance; other modifications, relying on the kernel density estimation methods attempt to reduce the unbalanced contribution of the different types of variables. The performance of the proposals is evaluated in simulations mimicking the imputation of missing values through nearest neighbor distance hotdeck method.

# 1. Introduction

Distances between observations have been extensively used in clustering applications where the objective is to group units being homogenous with respect to the studied phenomenon; nowadays these techniques fall under the umbrella of *unsupervised learning*, in the framework of *statistical learning techniques* (often also denoted as *machine learning*). In particular, a huge amount of literature on distance functions can be found in articles on *hierarchical clustering* (see e.g. Kaufman and Rousseeuw, 1990). Recently, distance-based methods have become popular also in the area of *supervised learning*, in both classification and regression problems. As noted by Hastie *et al.* (2009, p. 459), their popularity originates from their simplicity and the fact that avoid fitting statistical models. The methods are very effective in classification problems and in regression applications but when the objective is that of getting predictions and not understanding the nature of the relationship between the predictors and the response variables. Hastie *et al.* (2009) warn that when distance methods are applied with many predictors, the "bias–variance tradeoff does not work as favorably for nearest-neighbor regression as it does for classification".

Distance-based methods have become very popular in official statistics mostly in the imputation of missing values (cf. Andridge and Little, 2010) or in data integration (*deterministic record linkage* and *data fusion*; cf. Coli *et al*, 2016; D'Orazio *et al.*, 2006). In imputation problems a missing value is replaced with the value observed on the closest responding unit (nearest neighbor donor hotdeck); this way of working avoids fitting of models and replacing the missing values with "artificial" model predictions.

Although the distance methods appear simple and easy to apply compared to fitting a statistical model, their application and the corresponding results are strictly dependent on the two key choices: (i) which of the available variables should be used to calculate the distance and, (ii) the distance function to apply; the two choices are strictly related. The first decision depends on the scope of the application; typically, in an imputation problem these variables are the best predictors of the variable to impute (also being able to explain the nonresponse mechanism). D'Orazio *et al.* (2019) discuss variable selection in data fusion applications. In both the cases, the parsimony is the guiding principle: relatively few variables can ensure good results. More in general, for many distance measures the capacity of using the distance for discriminating between observations diminish as the number of variables increases, i.e. everything starts being close to everything; this effect is known as the *curse of dimensionality* (cf. Marimont and Shapiro, 1979). How to select the variables to be used in calculation of distance is not the scope of this work.

The choice of the distance function depends mainly on the type of the variables selected in the step (i). Here it is considered a well-known classification of the variables according to the measurement scale: *categorical nominal*, *categorical ordinal*, *measured on an interval scale*, *measured on a ratio scale*. A categorical nominal variable can assume a finite number of categories (if only two it is said binary variable) that do not have a natural ordering. The categories of a categorical ordinal variable have a natural ordering but the difference between categories is not meaningful. Variables measured on an interval-scale show quantitative measurement (integer or real numbers) and the difference between values is meaningful, but not the ratio (multiplication). Finally, a variable measured on a ratio scale is an interval-scaled variable that admits the zero and where the ratio is meaningful. Often the categorical variables are also identified as *qualitative* while the interval or ratio scale variables are said *quantitative* or *continuous*.

When all the variables are all categorical or all non-categorical, many distance functions can be applied. Unfortunately, relatively few options are available in the presence of mixed-type variables, a situation frequently encountered when working with data collected in a survey on households or single individuals. The most popular distance for mixed-type variables is derived as the complement of the *Gower's similarity coefficient* (Gower, 1971); an average of the distances calculated variable by variable, whereas the single distances are all scaled to range from 0 (minimum distance) to 1 (maximum distance). The Gower's distance is appealing because allows for missing values and permits to introduce a user-defined weighting scheme when averaging distances over the single variables. Unfortunately, the discussion on the weighting schemes is sometimes misleading as it often ignores that in absence of weights (unweighted "standard" version) the categorical variables have a higher impact on the final distance than the numeric ones, as it will be shown later.

Other distances functions for mixed-type variables have been introduced in clustering applications; most of them consist in a linear combination of the distances calculated on subgroups of variables of the same type. Some of the proposed distances can be viewed as extensions of the *K-means* (see e.g. van de Velden *et al.*, 2018; Ahmad and Khan, 2019). An issue that limits the applicability of these distances outside the clustering problems is the fact that the weights assigned to the different components of the overall distance are typically optimized taking into the characteristics of the clustering procedure; inn addition, in most of the cases, the weights are not constant but updated in the various clustering steps. Finally, many of the proposed distances do not permit to handle the missing values.

Simplest approaches for calculating distances with mixed-type distances consist in replacing each categorical variable with the corresponding dummies and treat them as quantitative or, in alternative, categorizing the interval and ratio-scaled variables. This latter choice implies a loss of information and the decision about the number of categories is not straightforward. Replacing a categorical nominal variable with the corresponding dummy variables increases the number of dimensions; secondly, the outcomes of the final distance depend on the function being adopted, as shown later.

In official statistics, when using distance-based methods in imputation or data integration problems, an alternative solution to handle mixed-type variables often consists in calculating "conditional distances". In practice, initial groups of units (*donors' pools*) are created by considering units sharing the same value(s) for the chosen categorical variable(s) then the distance between units in the same donation class is calculated using the remaining quantitative variables. The distance between units belonging to different groups is not calculated (see e.g. Andridge and Little, 2010) (practically this corresponds to set the distance between them to the maximum achievable value). This way of working gives implicitly a higher importance to categorical variables used in forming the donors' pools.

This work will focus on the Gower's distance because it has the advantage of being applicable for solving different problems (clustering, imputation, etc.). We address the main drawbacks that affect the calculation of the overall unweighted Gower's distance by suggesting some modifications that try to attenuate the impact of outliers in the ratio-scaled variables and to balance the contribution of the different variables to the overall distance. The remaining of the article is organized as follows. The Section 2 illustrates the main characteristics of the Gower's similarity. In particular, the Section 2.1 shows how the different composition of the set of the chosen variables contributes to the overall unweighted Gower's distance. The Section 3 introduces the proposed modifications of the standard Gower's distance to compensate for its drawbacks. The behavior of these modifications is investigated in two simulation studies whose results are presented in the Section 4. Finally, the Section 5 summarizes the main findings and the future areas of work.

## 2. The Gower's distance

The Gower's (1971) proposal is the most popular way of measuring the similarity/dissimilarity between observations in the presence of mixed-type variables. The *Gower's distance* can be defined as the complement to one of the Gower's *similarity coefficient*:

$$d_{G,ij} = 1 - s_{G,ij} = \frac{\sum_{t=1}^{p} \delta_{ijt}\, d_{ijt}}{\sum_{t=1}^{p} \delta_{ijt}} \qquad [1]$$

It is a *dissimilarity*[1] or *distance* measure (Kaufman and Rousseeuw, 1990, p. 35) between unit $i$ and unit $j$, where $d_{ijt} = 1 - s_{ijt}$ is the distance calculated on the $t$th variable; $s_{ijt}$ is the similarity between $i$ and $j$ with respect to the $t$th variable and its value depends on the type of the variable itself. If $x_{it}$ is the value observed for the $t$th variable ($t = 1,2,...,p$) on the $i$th unit ($i = 1,2,...,n$), the possible values of $s_{ijt}$ (and consequently of $d_{ijt} = 1 - s_{ijt}$) are reported in Table 1.

---

[1] The dissimilarity is a wider concept than that of *metric*; it satisfies only the first two properties of a metric ($d_{G,ij} \geq 0$) but not the triangle inequality (for more detailed discussion see e.g. Gower and Legendre, 1986).

Table 1 – Calculation of the Gower's similarity by type of variable

| Type of variable | $s_{ijt}$ | $\delta_{ijt}$ | Note |
|---|---|---|---|
| Binary symmetric | 1 if $x_{it} = x_{jt}$<br>0 if $x_{it} \neq x_{jt}$<br>0 if $x_{it}$ or $x_{jt}$ or both are missing | 1 if both the variables are nonmissing<br>0 if $x_{it}$ or $x_{jt}$ or both are missing | $s_{ijt}$ corresponds to the *simple matching coefficient* |
| Binary asymmetric | 1 if $x_{it} = x_{jt} = 1$<br>0 otherwise<br>0 if $x_{it}$ or $x_{jt}$ or both are missing | 1 if both the variables are nonmissing<br>0 if $x_{it} = x_{jt} = 0$<br>0 if $x_{it}$ or $x_{jt}$ or both are missing | $s_{ijt}$ corresponds to the *Jaccard index* |
| Categorical nominal (more than two categories) | 1 if $x_{it} = x_{jt}$<br>0 if $x_{it} \neq x_{jt}$<br>0 if $x_{it}$ or $x_{jt}$ or both are missing | 1 if both the variables are nonmissing<br>0 if $x_{it}$ or $x_{jt}$ or both are missing | $s_{ijt}$ is the simple matching on the untransformed variable or to the *Dice* (*Czekanovsky-Sorerensen*) *measure* applied to the dummies obtained by transforming the original variable |
| Measured on an interval or ratio scale | $1 - |x_{it} - x_{jt}|/R_t$<br>0 if $x_{it}$ or $x_{jt}$ or both are missing | 1 if both the variables are nonmissing<br>0 if $x_{it}$ or $x_{jt}$ or both are missing | $R_t = \max(x_t) - \min(x_t)$ is the range of the *k*th variable<br>$1 - s_{ijt}$ is *the Manhattan* or *city-block distance* scaled by the range |

The original Gower's proposal does not provide specific guidance for categorical ordered variables. Kaufman and Rousseeuw (1990) suggest to replace the categories of the variable with the corresponding position index in their natural ordering ($o_{it}; 1 \leq o_{it} \leq C$) and then derive a new variable:

$$z_{it} = \frac{o_{it}-1}{\max(o_{it})-1} \qquad [2]$$

that, for the calculation of the similarity, is treated as a variable measured on a ratio scale.

Podani (1999) replaces the ordered categories with the corresponding ranks ($r_{it}$) and then suggests to calculate the similarity as follows:

$$s_{ijt} = 1 - \frac{|r_{it}-r_{jt}|}{\max(r_{it})-\min(r_{it})} \qquad [3]$$

This latter expression requires a correction to account for ties (see Podani, 1999).

In practice, the Gower's dissimilarity ($d_{G,ij} = 1 - s_{G,ij}$) can be viewed as an average of the dissimilarities ($d_{ijt} = 1 - s_{ijt}$) measured on the available variables, where the dissimilarity calculated on each single variable is scaled to range from 0 (most dissimilar units) to 1 (maximum distance); as a consequence, the averaging will provide an overall dissimilarity taking values between 0 and 1 ($0 \leq d_{G,ij} \leq 1$).

The Gower's dissimilarity allows for missing values, which do not contribute to the calculation of the overall dissimilarity. Obviously, if a unit presents missing values for all the *p* variables then the dissimilarity with any other unit with partially observed data would be undefined; for this reason, a unit with all the values missing should be discarded before the calculations.

The Gower's proposal is the most popular approach for measuring similarity/dissimilarity in presence of mixed-type variables. As mentioned before, simplest approaches for calculating distances with mixed-type distances consist in replacing the categorical variable with the corresponding dummies and treat them as quantitative. This choice has many drawbacks since increases the number of dimensions; secondly, the achieved distances change according to the distance function adopted. In practice, the maximum value of the distance depends on the number the number $c$ of categories (dummies) of the starting categorical nominal variable. In general, when dealing with a set of solely categorical variables the following result holds:

$$d_{Dice} = \frac{d_{Manh}}{c-1} = \frac{d_{Euc}^2}{c-1} = \frac{p}{c-1} d_{SM} \qquad [4]$$

Where *c*, the number of dummies, is equal to all the possible combinations of the categories of the chosen $p_{cat}$ categorical variables. Unfortunately, *c* increases rapidly and consequently the risk of running into the curse of dimensionality issues (cf. Marimont and Shapiro, 1979).

A generalization of the Gower's dissimilarity consists in using a weighted average:

$$d_{wG,ij} = \frac{\sum_{t=1}^{p} \delta_{ijt} d_{ijt} w_t}{\sum_{t=1}^{p} \delta_{ijt} w_t} \qquad [5]$$

that assigns a different weight, $w_t$, to each of the variables (the unweighted version corresponds to setting $w_t = 1$ for all the variables). As remarked by Gower (1971), the decision on a rational set of weights is difficult. Generally speaking, the discussion about potential alternative set of weights is often flawed by the thinking that the unweighted Gower's coefficient, assigning an equal weight to each of the variables ($w_t = 1$), corresponds to a balanced contribution of the various variables to the overall distance. In practice, this is not completely true and, in addition, the presence of outliers in numerical variables influences their contribution to the final overall dissimilarity, as will be shown in the next sub-section.

In the R environment (R Core Team, 2020) there are several implementations of the Gower's distance/similarity. The most popular is the `daisy` function in the **cluster** package (Maechler *et al.*, 2019); the function `gowdis` in the package **FD** (Laliberté *et al.*, 2014) implements also the various options suggested by Podani for calculating the distance on categorical ordered variables. The package **gower** (van der Loo, 2020) allows to calculate the Gower's distance as well as the top-*n* matches between records; it is very efficient and fast permitting execution in parallel too. It is worth mentioning the R package **kmed** (Budiaji, 2019) that implements the Gower's and other distance functions for mixed-type variables, but the proposed functions do not handle categorical ordered variables nor the missing values.

*2.1 The contribution of different types of variables to the Gower's distance*

The unweighted Gower's distance assigns the same weight to each chosen variable and the final overall distance is just a simple average of distances calculated on each single variable. Since the distance on each single variable can vary between 0 (minimum distance) and 1 (maximum distance) also the final average will be included in this interval, and this feature is very appealing for practitioners. The mechanism that ensures such a result, is basically a standardization of the single distances so to have them ranging between 0 and 1; this operation is quite straightforward but can have a number of undesired consequences. Let's focus attention on the simple case of a categorical nominal variable and a quantitative one; for the categorical variable $d_{ijt} = 0$ for all the cases where $x_{it} = x_{jt}$, this situation will occur less frequently for the quantitative variable where $d_{ijt} = 0$ only when $|x_{it} - x_{jt}| = R_t$. In other words, the standard unweighted Gower's distance shows an unbalanced contribution of quantitative variables compared to categorical ones and the problem is not trivial since the final distance depends also on the number of variables by type, their distribution and the number of categories of the various categorical variables (see e.g. Foss *et al.*, 2016).

To better illustrate the problem let us consider the following example showing some possible values of the Gower's distance between two individuals according to various combinations of gender and age. The Table 2 reports the results ordered according to increasing values of the final distance (last column).

Table 2 – An example of calculation of Gower's distance

| Sex1 | Sex2 | d.Sex | Age1 | Age2 | d.age* | Gower |
|---|---|---|---|---|---|---|
| M | M | 0 | 15 | 15 | 0 | 0.0000 |
| M | M | 0 | 15 | 36 | 0.25 | 0.1235 |
| F | F | 0 | 15 | 58 | 0.51 | 0.2529 |
| F | F | 0 | 15 | 78 | 0.74 | 0.3706 |
| F | F | 0 | 15 | 100 | 1 | 0.5000 |
| M | F | 1 | 15 | 15 | 0 | 0.5000 |
| M | F | 1 | 15 | 36 | 0.25 | 0.6235 |
| F | M | 1 | 15 | 58 | 0.51 | 0.7529 |
| F | M | 1 | 15 | 78 | 0.74 | 0.8706 |
| F | M | 1 | 15 | 100 | 1 | 1.0000 |

* For calculating the distance on age ("d.age" column) it is considered R_age=|100-15|=85, assuming that the sample observes only individuals with age>14.

The Table 2 shows quite well that in identifying the closer units the Gower's distance tends to favor the observations sharing the same gender; two units having the same gender but a huge distance on age (15 vs 78, $d_{G,12} = 0.3706$) are in the end closer than two units having the same age but a different gender ($d_{G,12} = 0.50$). The latter case is equivalent to the situation of units sharing the same gender but with maximum distance on the age (15 vs 100, $d_{G,12} = 0.50$). The example shows clearly that $d_{age,12} = 1$ is achieved only when comparing the two extreme values in the opposite tails of the age distribution (as observed on the available sample), where the value of 100 can be considered quite rare. This reveals an additional problem of the range-scaled Manhattan distance: the presence of outliers in the distribution of the continuous variable may affect directly the estimation $R_t$ and, as a consequence, the values of $d_{ijt}$ become smaller ($d_{ijt} = 1$ would be observed only rarely).

The unbalanced contribution of the different type of variables is further exacerbated when the number of categorical variables increases. If, for instance, the marital status is considered in addition to gender and age, then units sharing the same gender and marital status will have a distance smaller than that of two individuals with a close age but a different gender or marital status. In practice, the Gower distance tend to consider as closer the units sharing the same values of the categorical variable, caring less on their distance on the ratio-scaled variable. This behavior can only be modified by adopting a non-uniform weighting scheme so to assign a higher weight $w_t$ to the interval or ratio-scaled variables. The choice of the weight would however remain highly subjective and not simple at all in the case of several mixed-type variables; moreover, adding or removing one variable would require resetting the whole system of weights. For these reasons, this paper will focus on possible modification of the unweighted Gower's coefficient instead of proposing a way of defining unequal weights. Basically, the idea is to modify the way of measuring the distance for the variables measured on an interval or ratio scale.

## 3. Possible modifications of the unweighted Gower's dissimilarity

A first modification of the standard Gower's distance to compensate for the impact of outliers in interval or ratio-scaled variables consists simply in the replacement of the range $R_t$ with the *inter-quartile* range, by setting

$$d_{ijt} = \begin{cases} \frac{|x_{it} - x_{jt}|}{IQR_t}, & \text{if } |x_{it} - x_{jt}| < IQR_t \\ 1, & \text{otherwise} \end{cases} \qquad [6]$$

where $IQR_t$ is the inter-Quartile range ($P75\% - P25\%$) (the *inter-decile* can also be considered). This modification permits to account for outliers but does not solve the problem of "dominance" of categorical variables on the overall distance. This dominance can partly be tackled by discretizing the interval and ratio-scaled variables. This work will avoid using a "fixed" discretization scheme since its results can vary markedly according to the chosen categorization criterion (equal-width classes, equal-frequency classes, etc.) and the decided number of categories. In alternative, a non-fixed discretization is proposed; following the rationale of kernel density estimation, two observations falling within the same moving "window" will be considered being at distance 0, while the distance for a couple of units where one or both of them do not fall in the window will remain calculated in the usual setting:

$$d_{ijt}^{(kde)} = \begin{cases} 0, & \text{if } |x_{it} - x_{jt}| \leq h_t \\ \frac{|x_{it} - x_{jt}|}{g_t}, & \text{if } h_t < |x_{it} - x_{jt}| < g_t \\ 1, & |x_{it} - x_{jt}| \geq g_t \end{cases} \quad [7]$$

In this expression $g_t$ can be the range (standard Gower's distance) or the IQR, while $h_t$ is the *window width* (*bandwidth* in the kernel density estimation) and plays a key role; for the sake of simplicity, we consider the "basic" Silverman (1986, pp. 45-48) suggestion:

$$h_t = c \frac{1}{n^{1/5}} \min\left\{s_t, \frac{IQR_t}{1.34}\right\} \quad [8]$$

where $s_t$ is the estimated standard deviation for the *t*th variable. According to Silverman (1986, p. 48) $c = 1.06$ performs quite well with skewed unimodal distributions while $c = 0.9$ should work better with skewed and moderately bimodal distributions. More sophisticated methods for choosing the bandwidth ($h_t$) will not be explored since this is beyond the objectives of the work.

The expression [7] avoids the major drawback of fixed discretization, i.e. that units belonging to two adjacent classes can actually be closer than units belonging to the same class but being at the opposite boundaries. For instance, when using equal-width age classes two units with respectively the maximum value of one class and the minimum value of the subsequent class, often have a smaller distance than units with values at the opposite extremes of the same class.

Following the same philosophy of kernel density estimator, another possible modification of the distance for interval and ratio scale variables consists in applying the *k*-nearest neighbor (*k*-nn) method:

$$d_{ijt}^{(knn)} = \begin{cases} 0, & \text{if } x_{jt} \text{ is one of the } k \text{ nearest neighbors of } x_{it} \\ \frac{|x_{it} - x_{jt}|}{g_t}, & x_{jt} \text{ not one of } k \text{ nearest neighbors of } x_{it} \text{ AND } |x_{it} - x_{jt}| < g_t \\ 1, & \text{if } |x_{it} - x_{jt}| \geq g_t \end{cases} \quad [9]$$

For the purposes of this work, it will be considered the well-known rule of thumb $k = \sqrt{n}$. The *k*nn-based method is known to perform better than kernel density in the tails of the distribution and for this reason $d_{ijt}^{(knn)}$ may be preferable to $d_{ijt}^{(kde)}$ when measuring distance on continuous variables with a highly skewed distribution with a long tail.

In addition to the previous suggested modifications of the way of calculating the distance on interval and ratio scale variables in the context of the Gower's philosophy, an alternative proposal of this work consists in a two-stage approach that follows the rationale of the "conditional distances". As mentioned in Section 1, in official statistics it is common to calculate the distances only between observations in the same donors' pool, i.e. sharing the same category of the given categorical variable (or combination of categories of two or more categorical variables). Following this rationale, we propose a two-step approach:

1) Calculate distance using the Gower's formula only on the subset of $p_{cat}$ variables being binary or categorical nominal.

2) Calculate the distance using the interval or ratio scaled variables by using the Gower approach (Manhattan distance scaled by the range or the IQR) only between observations whose distance in step (1) is smaller or equal to $1/p_{cat}$ (if there are no units satisfying the condition then the threshold is increased to $2/p_{cat}$ and so on).

The final overall distance between two observations is set equal to Manhattan distance scaled by the range or by the IQR for units satisfying the condition on the distance on the subset of categorical variables, while for the remaining ones the distance is set to 1, the maximum value. This approach corresponds to calculate the distance on the interval or ratio-scaled variables solely between the observations that show a difference in one (or two) of the selected categorical variables. This is also very efficient from computational purposes since reduces the effort in calculation of overall distances.

## 4. An application in donor-based imputation

This Section will present the results of a series of simulations mimicking a donor-based imputation setting where the distance between respondents (donors) and non-respondent units (recipient) is calculated using the standard unweighted Gower's distance and its modification proposed in the Section 3. A first series of simulations is carried out on artificially generated data from a multivariate Gaussian distribution; a second series of simulations is based on real data related to income and consumption of the Italian households in 2016, as observed in the sample survey carried out by the Bank of Italy (Banca d'Italia, 2018).

*4.1. Simulations with artificial data*

The simulations carried out with artificial data consider a very "basic" framework. Each simulation consists of the following steps:

A1) a reference sample of data ($n = 500$) is generated from a multivariate Gaussian distribution with $p = 6$ variables sharing the same mean, $\mu = 100$, and standard deviation, $\sigma = 20$; while the correlation matrix is:

|       | $X_1$ | $X_2$ | $X_3$ | $X_4$ | $X_5$ |
|-------|-------|-------|-------|-------|-------|
| $Y$   | 0.8   | 0.4   | 0.8   | 0.4   | 0.5   |
| $X_1$ |       | 0.2   | 0.4   | 0.2   | 0.3   |
| $X_2$ |       |       | 0.2   | 0.2   | 0.3   |
| $X_3$ |       |       |       | 0.2   | 0.2   |
| $X_4$ |       |       |       |       | 0.2   |

A2) categorization of some variables, two options are considered:

    A2.1) the variables $X_2, \ldots, X_5$ are categorized in respectively 4, 6, 2 and 4 equal-width classes.

    A2.2) the variables $X_3$, $X_4$ and $X_5$ are categorized in respectively 6, 2 and 4 equal-width classes.

A3) perturbation of $X_1$. Two options:

    3.1) No changes of the starting values;

    3.2) Introduction of outliers: about 2% of randomly chosen observations have the values of $X_1$ replaced with those generated form a Gaussian distribution with $\mu_o = 2\mu = 200$ and $\sigma_0 = \sigma = 20$.

A4) About 1/3 of the values of $Y$ are randomly deleted and then imputed using the nearest neighbor hotdeck method; the distance between recipients and donors is calculated on the variables $X_1, \ldots, X_5$ by means of the standard Gower's distance and its variants introduced in Section 3. In particular, both scaling by range ($g_t = R_t$) and by IQR ($g_t = IQR_t$) are considered.

The whole procedure is run 1,000 times for each combination of the various options for steps (A2) and (A3); the final results are evaluated by measuring the closeness between the initial deleted values with the corresponding ones imputed through the Pearson's correlation coefficient. Then, it is assessed how imputed and observed values are able to reproduce the true mean:

$$sB = \frac{1}{1000}\sum_{h=1}^{1000}[\bar{y}_{obs\&imp,h} - \mu_Y]; \quad sRMSE = \sqrt{\frac{1}{1000}\sum_{h=1}^{1000}(\bar{y}_{obs\&imp,h} - \mu_Y)^2}$$

as well as their ability in reproducing the "true" distribution of $Y$; this is assessed by calculating the average differences between the percentiles estimated using imputed and observed values ($\hat{q}_{obs\&imp,k}$) and the theoretic percentiles $q_k$ (2.5% percentiles are considered); both average of differences and of square root of squared differences are calculated:

$$sDQ = \frac{1}{1000}\sum_{h=1}^{1000}\left[\frac{1}{41}\sum_{k=1}^{41}(\hat{q}_{obs\&imp,hk} - q_k)\right]; \quad sRSDQ = \frac{1}{1000}\sum_{h=1}^{1000}\sqrt{\frac{1}{41}\sum_{k=1}^{41}(\hat{q}_{obs\&imp,hk} - q_k)^2}$$

The results of the various simulations are summarized in the Table 3.

Table 3 – Results of the simulation study on artificially generated data

| Case | Assess criterion | Scaling by the range | | | | | Scaling by IQR | | | | |
|---|---|---|---|---|---|---|---|---|---|---|---|
| | | no.mod | kde1 | kde2 | knn | cond.dist | no.mod | kde1 | kde2 | knn | cond.dist |
| 1 cont. 4 cat. No outl. | $\rho_{imp,true}$ | 0.8913 | 0.8894 | 0.8906 | 0.8905 | 0.8167 | 0.8831 | 0.8895 | 0.8900 | 0.8924 | 0.8145 |
| | sB | -80.3505 | -80.3505 | -80.3466 | -80.3600 | -80.2787 | -80.3446 | -80.3749 | -80.3761 | -80.3379 | -80.2829 |
| | sRMSE | 80.3536 | 80.3535 | 80.3497 | 80.3630 | 80.2818 | 80.3477 | 80.3780 | 80.3791 | 80.3409 | 80.2860 |
| | sDQ | -0.0174 | -0.0167 | -0.0223 | -0.0191 | 0.0052 | -0.0160 | -0.0181 | -0.0187 | -0.0183 | 0.0045 |
| | sRSDQ | 99.9399 | 99.9403 | 99.9363 | 99.9377 | 99.9623 | 99.9409 | 99.9407 | 99.9408 | 99.9396 | 99.9613 |
| 1 cont. 4 cat. With Outl | $\rho_{imp,true}$ | 0.8840 | 0.8821 | 0.8827 | 0.8833 | 0.8050 | 0.8800 | 0.8851 | 0.8858 | 0.8829 | 0.8079 |
| | sB | -80.2681 | -80.2669 | -80.2662 | -80.2760 | -80.1971 | -80.2764 | -80.3125 | -80.3144 | -80.2511 | -80.2295 |
| | sRMSE | 80.2713 | 80.2700 | 80.2694 | 80.2792 | 80.2002 | 80.2795 | 80.3156 | 80.3174 | 80.2541 | 80.2326 |
| | sDQ | 0.0501 | 0.0481 | 0.0498 | 0.0526 | 0.0546 | -0.0061 | -0.0037 | -0.0049 | 0.0335 | 0.0056 |
| | sRSDQ | 100.0569 | 100.0542 | 100.0558 | 100.0595 | 100.0635 | 99.9992 | 100.0004 | 99.9995 | 100.0414 | 100.0103 |
| 2 cont. 3 cat. No outl. | $\rho_{imp,true}$ | 0.9135 | 0.9125 | 0.9125 | 0.9132 | 0.7951 | 0.8990 | 0.8971 | 0.8966 | 0.9021 | 0.7854 |
| | sB | -80.2618 | -80.2639 | -80.2636 | -80.2511 | -80.2784 | -80.3393 | -80.3613 | -80.3650 | -80.2961 | -80.3011 |
| | sRMSE | 80.2647 | 80.2668 | 80.2665 | 80.2540 | 80.2815 | 80.3421 | 80.3641 | 80.3679 | 80.2989 | 80.3043 |
| | sDQ | -0.0022 | 0.0011 | -0.0016 | 0.0009 | 0.0101 | -0.0069 | -0.0013 | -0.0061 | -0.0012 | 0.0083 |
| | sRSDQ | 100.0095 | 100.0118 | 100.0097 | 100.0119 | 100.0168 | 100.0041 | 100.0079 | 100.0037 | 100.0085 | 100.0162 |
| 2 cont. 3 cat. With Outl | $\rho_{imp,true}$ | 0.8961 | 0.8940 | 0.8939 | 0.8941 | 0.7758 | 0.8955 | 0.8930 | 0.8925 | 0.8923 | 0.7751 |
| | sB | -80.2577 | -80.2638 | -80.2635 | -80.2377 | -80.2958 | -80.3314 | -80.3518 | -80.3492 | -80.2509 | -80.3038 |
| | sRMSE | 80.2606 | 80.2667 | 80.2664 | 80.2407 | 80.2989 | 80.3343 | 80.3546 | 80.3521 | 80.2538 | 80.3069 |
| | sDQ | 0.0598 | 0.0554 | 0.0568 | 0.0692 | 0.0502 | 0.0121 | 0.0082 | 0.0109 | 0.0618 | 0.0131 |
| | sRSDQ | 100.0649 | 100.0617 | 100.0630 | 100.0752 | 100.0538 | 100.0128 | 100.0095 | 100.0144 | 100.0659 | 100.0154 |

In general, scaling the Manhattan distance by the IQR ($g_t = IQR_t$) tends to perform better than scaling by range ($g_t = R_t$) when $X_1$ shows some outliers, in particular in preserving the true distribution (criteria "sDQ" and "sRSDQ"). When the number of continuous variables increases (2 instead of 1), the modified distances based on the *k*-nn and kernel density estimation ("k-nn", "kde1"and "kde2") seem to perform slightly better in preserving the "true" deleted values and the marginal distribution of the target variable. As expected, scaling by the IQR gives slightly better results in preserving the marginal distribution of *Y* in the presence of outliers. Generally speaking, the results are very close when the focus is on estimating the mean of the distribution; the modification of Gower's distance based on the ideas underlying the "conditional distance" (column "cond.dist") seems to perform slightly better when there is just one continuous variable. However, this proposal does not perform well in reproducing the "true" deleted values, where shows always the worst performances.

4.2 Simulations carried out on real survey data

This subsection reports the summary results of a series of simulations carried out on using the data observed in the Survey of Household Income and Wealth in Italy related to year 2016, a sample survey carried out every two years by the Bank of Italy (2018). These simulations are important to understand how the proposed modifications of the Gower's distance behave with real observed data when the variable to impute as well some of the available ratio-scaled predictors show a right-skewed distribution, typical of income and consumption related HH variables.

The anonymized microdata[2] related to the about 8,000 Italian households (HHs) are used as basis for the simulations; in particular, each simulation run consists of the following steps:

> B1) a subsample of 1,000 HHs is selected out of the whole sample; selection is done with probability proportional to the final HH survey weight;
>
> B2) about 1/3 of the subsample values of the HH consumption (*C*) are deleted following different mechanisms:
>
>> B2.1) nonrespondents generated completely at random (MCAR)
>>
>> B2.2) nonrespondents generated with probability proportional to the HH net disposable income (*Y*) (MAR)
>>
>> B2.3) nonrespondents generated with probability proportional to the HH consumption (*C*) (MNAR)
>
> B3) the missing values on *C* are imputed by means of the nearest neighbor method; the distance between the recipients and donors is calculated considering two sets of predictors, as shown in Table 4, by applying the standard Gower's distance and its variants introduced in Section 3. Both scaling by range ($g_t = R_t$) and by IQR ($g_t = IQR_t$) are considered.
>
>> The last column in the Table 4 shows the adjusted $R^2$ related to the linear regression model fitted by considering each single variable (replaced by dummies for categorical variables or ranks for interval and ratio-scaled variable) when used as predictor of the ranks of the target one (*C*) (cf. Harrell, 2015, p. 460). This can be considered as an indication of the potential predictive power of each of the available variables.

---

[2] The anonymized microdata can be downloaded at the following location
https://www.bancaditalia.it/statistiche/tematiche/indagini-famiglie-imprese/bilanci-famiglie/distribuzione-microdati/index.html?com.dotmarketing.htmlpage.language=1

Table 4 – Household variables used to calculate the distance

| Predictors variable | Type | Set 1 | Set 2 | Adj. $R^2$ |
|---|---|---|---|---|
| HH net disposable income (Y) | Ratio scale | ✓ | ✓ | 0.61 |
| No. of HH income earners (NPERC) | Integer (1,2,3,…) | ✓ | ✓ | 0.22 |
| Education level of the major income earner in the HH (STUDIO) | Categorical (8 categories) | ✓ | ✓ | 0.19 |
| Employment status of the major income earner in the HH (QUAL) | Categorical (7 categories) | ✓ | ✓ | 0.14 |
| No. of HH members (NCOMP) | Integer (1,2, …) |  | ✓ | 0.14 |
| Marriage status of the major income earner in the HH (STACIV) | Categorical (4 categories) |  | ✓ | 0.12 |

The whole procedure is run 500 times for each combination of the various options for the steps (B2) and (B3); the performances are evaluated by considering the same criteria used for simulations on artificially generated data.

The results of simulations on the real survey data show a picture slightly different from that obtained with artificial data (see Tale 5). The modification Gower's distance relying on the conditional distance approach performs better for all the considered criteria, in particular when scaling in the Manhattan distance by the range. Slight inferior performances are shown by the modification based on *k*-nn, but with scaling based on IQR instead of the range. This is not surprising since some of the continuous variables used in calculating the distance have a marked right-skewed distribution which favors the *k*-nn to kernel density estimators. The simulations show that calculating the distance on the smallest set of available variables (set1) gives result better than those based on the larger one (set2). This confirms the need of maintaining the parsimony in selecting the variables to be used in calculation of the distance.

## 5. Conclusions

This article suggests some modifications of the standard unweighted Gower's dissimilarity index needed to calculate the distance in the presence of with mixed-type variables. In particular, it is proposed a series of alternative modifications in the way of calculating the distance for the interval or ratio-scaled variables. The proposed modifications try to account for the presence of outliers in the quantitative variables, for the skewness in the distribution and attempt to better balance the contribution of the different types of variables on the overall unweighted Gower's distance. The results obtained in the simulation studies seem to suggest that IQR-based scaling should be preferred when calculating the distance on the interval and ratio scale variables in presence of outliers. The modification following the *k*-nn ideas seems to work better when the continuous variables show a high skewed distribution, typically shown by economic variables observed in household surveys or enterprises surveys. The conditional approach seems the more effective in reducing the unbalanced contribution of the different types of variables in the overall distance when the continuous variables show a skewed distribution.

The proposed modifications of the Gower's seem to provide promising results when used in imputation problems, however further investigation is need to understand their behavior in different applications in absence of a target variable, i.e. in unsupervised statistical learning.

Table 5 – Results of the simulation study on HH income and consumption data

| Case | Assess criterion | Scaling by the Range | | | | | Scaling by IQR | | | | |
|---|---|---|---|---|---|---|---|---|---|---|---|
| | | no.mod | kde1 | kde2 | knn | cnd.dist | no.mod | kde1 | kde2 | knn | cnd.dist |
| MCAR set1 | $\rho_{imp,true}$ | 0.5648 | 0.5577 | 0.5602 | 0.5659 | 0.6020 | 0.5383 | 0.5425 | 0.5413 | 0.5654 | 0.5636 |
| | sB | -111.30 | -122.63 | -124.84 | -128.24 | -62.19 | -130.87 | -147.76 | -163.53 | -102.41 | -122.94 |
| | sRMSE | 17.52 | 17.59 | 17.57 | 17.65 | 15.97 | 18.04 | 17.67 | 18.15 | 17.36 | 16.89 |
| | sDQ | -346.18 | -358.10 | -358.31 | -353.34 | -300.65 | -371.14 | -386.37 | -397.51 | -342.36 | -350.64 |
| | sRSDQ | 2428.74 | 2423.62 | 2424.52 | 2433.35 | 2408.48 | 2434.31 | 2421.58 | 2424.23 | 2427.71 | 2407.72 |
| MCAR set2 | $\rho_{imp,true}$ | 0.5059 | 0.5056 | 0.5055 | 0.5139 | 0.5881 | 0.5163 | 0.5285 | 0.5293 | 0.5458 | 0.5439 |
| | sB | -146.74 | -150.61 | -151.63 | -144.19 | -85.96 | -127.91 | -149.98 | -152.94 | -93.81 | -188.22 |
| | sRMSE | 17.61 | 17.70 | 17.71 | 17.54 | 16.04 | 17.09 | 17.33 | 17.43 | 16.58 | 17.74 |
| | sDQ | -368.16 | -373.69 | -372.98 | -362.61 | -297.88 | -357.76 | -371.08 | -375.69 | -319.61 | -388.93 |
| | sRSDQ | 2365.99 | 2366.51 | 2364.88 | 2367.58 | 2339.63 | 2353.81 | 2348.72 | 2356.22 | 2345.71 | 2356.98 |
| MAR set1 | $\rho_{imp,true}$ | 0.3885 | 0.3883 | 0.3888 | 0.4118 | 0.4835 | 0.3121 | 0.3142 | 0.3133 | 0.4058 | 0.2928 |
| | sB | -949.75 | -976.65 | -972.48 | -894.58 | -423.43 | -1077.10 | -1056.72 | -1072.14 | -785.03 | -914.13 |
| | sRMSE | 32.20 | 32.58 | 32.51 | 31.39 | 25.57 | 33.84 | 33.54 | 33.75 | 29.63 | 31.66 |
| | sDQ | -2532.84 | -2560.47 | -2555.09 | -2491.23 | -2060.11 | -2657.54 | -2643.14 | -2652.35 | -2395.55 | -2506.29 |
| | sRSDQ | 12673.34 | 12674.92 | 12676.04 | 12661.46 | 12627.10 | 12690.48 | 12682.63 | 12687.37 | 12641.56 | 12646.78 |
| MAR set2 | $\rho_{imp,true}$ | 0.3477 | 0.3482 | 0.3483 | 0.3771 | 0.4348 | 0.3092 | 0.3156 | 0.3139 | 0.4071 | 0.2969 |
| | sB | -1292.67 | -1299.78 | -1300.65 | -1208.87 | -749.06 | -1205.59 | -1181.67 | -1188.83 | -856.34 | -1174.22 |
| | sRMSE | 36.92 | 37.02 | 37.02 | 35.85 | 29.31 | 35.70 | 35.36 | 35.49 | 31.21 | 35.06 |
| | sDQ | -2873.49 | -2882.98 | -2882.61 | -2808.61 | -2365.69 | -2792.86 | -2773.88 | -2777.24 | -2484.88 | -2753.81 |
| | sRSDQ | 12825.23 | 12825.85 | 12826.48 | 12806.68 | 12746.67 | 12812.64 | 12805.54 | 12808.23 | 12756.38 | 12808.43 |
| MNAR set1 | $\rho_{imp,true}$ | 0.4569 | 0.4572 | 0.4573 | 0.4712 | 0.5262 | 0.3804 | 0.3892 | 0.3884 | 0.4504 | 0.4051 |
| | sB | -2211.53 | -2247.32 | -2230.42 | -2226.89 | -1945.98 | -2311.41 | -2321.07 | -2316.51 | -2198.13 | -2143.76 |
| | sRMSE | 47.24 | 47.62 | 47.43 | 47.40 | 44.41 | 48.27 | 48.36 | 48.32 | 47.09 | 46.54 |
| | sDQ | -4163.76 | -4196.30 | -4179.94 | -4179.08 | -3916.53 | -4257.17 | -4268.98 | -4262.85 | -4151.46 | -4099.91 |
| | sRSDQ | 16346.76 | 16351.92 | 16347.91 | 16352.82 | 16280.65 | 16372.56 | 16373.77 | 16372.10 | 16343.17 | 16325.43 |
| MNAR set2 | $\rho_{imp,true}$ | 0.4051 | 0.4067 | 0.4060 | 0.4175 | 0.5035 | 0.3715 | 0.3775 | 0.3776 | 0.4389 | 0.3924 |
| | sB | -2427.57 | -2442.47 | -2431.12 | -2409.62 | -2106.90 | -2398.41 | -2388.49 | -2395.14 | -2243.60 | -2322.84 |
| | sRMSE | 49.46 | 49.61 | 49.50 | 49.28 | 46.15 | 49.15 | 49.04 | 49.12 | 47.57 | 48.37 |
| | sDQ | -4325.67 | -4339.11 | -4329.03 | -4309.72 | -4020.41 | -4297.44 | -4287.70 | -4292.27 | -4151.03 | -4223.38 |
| | sRSDQ | 16157.36 | 16158.03 | 16156.18 | 16153.32 | 16087.92 | 16154.22 | 16148.42 | 16151.25 | 16108.05 | 16138.89 |